# Training LSTM Networks with Resistive Cross-Point Devices

**Authors:** Tayfun Gokmen,* Malte Rasch and Wilfried Haensch

**Affiliations**
IBM Research AI, Yorktown Heights, NY 10598 USA
*Correspondence to: tgokmen@us.ibm.com

**Abstract**
In our previous work we have shown that resistive cross point devices, so called Resistive Processing Unit (RPU) devices, can provide significant power and speed benefits when training deep fully connected networks as well as convolutional neural networks. In this work, we further extend the RPU concept for training recurrent neural networks (RNNs) namely LSTMs. We show that the mapping of recurrent layers is very similar to the mapping of fully connected layers and therefore the RPU concept can potentially provide large acceleration factors for RNNs as well. In addition, we study the effect of various device imperfections and system parameters on training performance. Symmetry of updates becomes even more crucial for RNNs; already a few percent asymmetry results in an increase in the test error compared to the ideal case trained with floating point numbers. Furthermore, the input signal resolution to device arrays needs to be at least 7 bits for successful training. However, we show that a stochastic rounding scheme can reduce the input signal resolution back to 5 bits. Further, we find that RPU device variations and hardware noise are enough to mitigate overfitting, so that there is less need for using dropout. We note that the models trained here are roughly 1500 times larger than the fully connected network trained on MNIST dataset in terms of the total number of multiplication and summation operations performed per epoch. Thus, here we attempt to study the validity of the RPU approach for large scale networks.

# INTRODUCTION

Deep neural networks (DNN) [1] have made tremendous improvements in the past few years tackling challenging problems such as speech recognition [2] [3], natural language processing [4] [5], image classification [6] [7], and machine translation [8]. These accomplishments became possible thanks to advances in computing resources, availability of large amounts of data and clever choices of neural network architectures. For instance, the spatial correlation in the data are tackled by convolution neural networks (CNNs) [6] [9] [10] whereas the temporal correlations can be handled by recurrent networks [11]. One of the most common recurrent architectures is long short-term memory (LSTM) [12] [13]. LSTMs in combination with CNNs provide end-to-end trainable building blocks for composing complex neural network architectures, that are used for challenging tasks such as image captioning [14]. Training these large complex DNNs is extremely computational intensive task and today most of these workloads run on general purpose digital hardware such as CPUs and GPUs in a massively parallel fashion [15] [16] [17] [18]. There are many attempts to accelerate training of large scale DNNs by designing and using specialized digital hardware [19], such as Google's TPU [20] or Intel's KNL [21], relying on optimized multiplication and summation operations. In addition to the digital approaches, resistive cross-point device arrays are also promising candidates that perform the multiplication and summation operations in the analog domain which can provide massive acceleration and power benefits [22].

The concept of using resistive cross-point device arrays [22] [23] [24] [25] [26] [27] [28] as DNN accelerators has been tested, to some extent, by performing simulations for the specific case of fully connected [22] and convolutional neural networks [29]. The effect of various device properties and system parameters on training performance has been evaluated to derive the required device and system level specifications for a successful implementation of an accelerator chip to efficiently train DNNs [22] [30]. A key requirement is that these analog resistive devices must change conductance symmetrically when subjected to positive or negative pulse stimuli. Indeed, these requirements differ significantly from those needed for memory elements and therefore require either an additional circuit overhead [31] or systematic search for new physical mechanisms, materials and device designs to realize such an ideal resistive element for DNN training. In addition to these critical device properties, the peripheral circuits and the whole system needs to be designed carefully within the specifications for successful DNN training. For instance, the input number normalization and output signal bound detection are shown to be critical while training CNNs [29] on the MNIST dataset and therefore these techniques may be incorporated into the system design.

It is clear that resistive cross-point devices, so called resistive processing unit (RPU) device arrays, are a promising candidate for compute intensive DNN tasks; however, any future hardware that is targeting DNN applications needs to be able to offer solutions for handling a range of network architectures including fully connected, convolutional as well as recurrent layers. Here, we extend the application space of RPUs to recurrent neural networks. We show how to map the complex recurrent LSTM blocks to RPU arrays and test the effect of various device level imperfections and peripheral circuit constraints, such as input signal resolution, to the training accuracy of LSTM networks on a character based language model. We also study the effect of dropout during training LSTMs in the presence of device imperfections and system level constraints. Although dropout is critical while training large LSTMs with floating point numbers to mitigate overfitting, it turns out that for RPU simulations training is not significantly affected by dropout. This suggests that some of the device imperfections and noise in the hardware may act as a regularization term during training. However, we also show that among all device variations symmetric updates become increasingly more important and the asymmetry term needs to be minimized for successful training. Our results further emphasize the importance of device symmetry when realizing a resistive element suitable for DNN training.

**MATERIALS AND METHODS**

**LSTM Block**

The dynamics of an LSTM block [12] is described using deterministic transitions from previous state to current state as shown by equations below and with its corresponding computational graph in Figure 1:

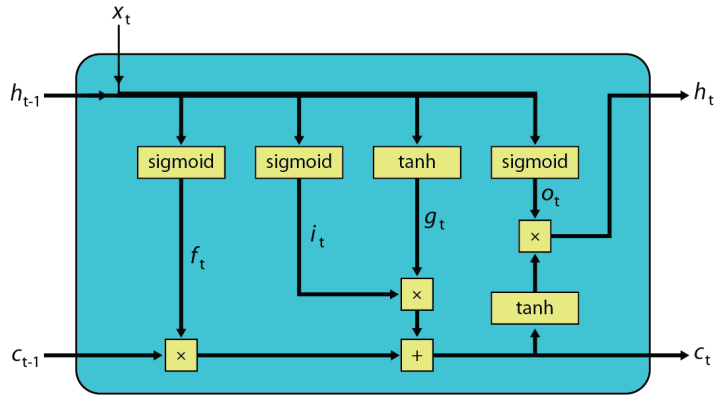

**Figure 1.** Computational graph of a LSTM block

$$f_t = sigmoid(W_f x_t + U_f h_{t-1} + b_f) \quad (1)$$
$$i_t = sigmoid(W_i x_t + U_i h_{t-1} + b_i) \quad (2)$$
$$o_t = sigmoid(W_o x_t + U_o h_{t-1} + b_o) \quad (3)$$
$$g_t = tanh(W_g x_t + U_g h_{t-1} + b_g) \quad (4)$$
$$c_t = f_t \times c_{t-1} + i_t \times g_t \quad (5)$$
$$h_t = o_t \times tanh(c_t) \quad (6)$$

where $x_t$ is the input vector of length $n$ for the current time step $t$, $h_{t-1}$ and $h_t$ are the hidden state vectors, $c_{t-1}$ and $c_t$ are the memory state vectors of length $m$ from the previous and current time steps, respectively. The trainable parameters for the LSTM block are stored in $W_f, W_i, W_o, W_c$ matrixes of sizes $m \times n$, $U_f, U_i, U_o, U_c$ matrixes of sizes $m \times m$ and bias terms $b_f, b_i, b_o, b_c$ of sizes $m \times 1$. $f_t, i_t, o_t$ and $g_t$ respectively correspond to the forget gate, input gate, output gate and new candidate memory state, all of which are vectors of length $m$. In these equations $sigmoid$ and $tanh$ functions are applied element-wise and $\times$ is an element-wise multiplication (Hadamard product).

Just like regular feed forward networks, LSTM networks are trained using backpropagation algorithm [32]. However, the concept of time in the case of an LSTM is simply expressed by a well-defined ordered series of calculations linking one time step to the next one and therefore error signals are backpropagated in time. The number of unrolling steps through time ($bptt$) used for backpropagation is a hyperparameter of the LSTM training. During training, all activations calculated during the forward pass for each time step need to be stored for the backward pass (for the derivative calculations). Once the backward pass is completed for all $bptt$ time steps, the total weight change, which is the sum of the gradients from each time step, can be calculated and applied to update the weights. Similar to the weight sharing concept for convolutional layers at different spatial locations, for an LSTM block the weights are shared between different time steps and the amount of sharing is controlled by the choice of $bptt$ during training.

**Mapping of an LSTM Block to Resistive Device Arrays**

Figure 2 illustrates all the calculations that need to be performed for an LSTM block during a forward pass and their mapping onto an RPU array. All of the trainable parameters of an LSTM block can be organized into a single matrix $W$ of size $4m \times (m + n + 1)$ which is then mapped onto a single RPU array of the same size. The temporary input vector to the RPU array that is used for each time step is shown as $\tilde{x}$ which is the concatenated vector of the input vector $x_t$ from current time step, the hidden state vector $h_{t-1}$ from the previous time step and a single bias value of unity. Performing a single vector matrix multiplication $\tilde{y} = W\tilde{x}$ yields a vector $\tilde{y}$ of length $4m$ where different portions can be used to calculate activations given by Eqs. (1)-(4) for a single time step. We note that the single $\tilde{y} = W\tilde{x}$ operation completes all the linear transformations that are needed and that has the computational complexity of $O(4m \times (m + n + 1))$. It can be performed with $O(1)$ time complexity once mapped to RPU arrays thanks to the array parallelism. All other computations shown above are point wise operations and therefore have the computational complexity of $O(m) + O(n)$. We assume this part of the computations are performed outside of the RPU array by a digital block, the so-called non-linear function unit (NLF). We note that all steps shown by Eqs. (1)-(6) are repeated $bptt$ times before error backpropagation starts; and similar computation steps are performed during the error backpropagation starting from the last time step. For instance, in the backward pass the computations that are performed on the array can be written as $\tilde{z} = W^T\tilde{\delta}$, where $\tilde{\delta}$ is the temporary error signal generated at each time step and $W^T$ is the transpose of the original weight matrix used during the forward pass. $\tilde{z}$ vector is further processed by the NLF units so that the error signal for the previous time step can be generated. Once the backward steps are repeated $bptt$ times, the weight update can be written as a series of updates $W \leftarrow W + \eta(\tilde{\delta}\tilde{x}^T)$ that is again performed $bptt$ times for the $\tilde{x}$ and $\tilde{\delta}$ vectors used at each iteration step during the forward and backward pass. Therefore, a LSTM block can be viewed as a fully connected layer but with parameter sharing that happens between time steps by reusing the same weight matrix for each step of the calculation. We emphasize that all other non-linear operations of the LSTM block are performed by the NLF units outside the array; and these NLF units require an access to a local or an external storage (memory) in order to save the intermediate results computed during the forward (and backward) pass that are also needed during the error calculation at backward pass (and update cycle).

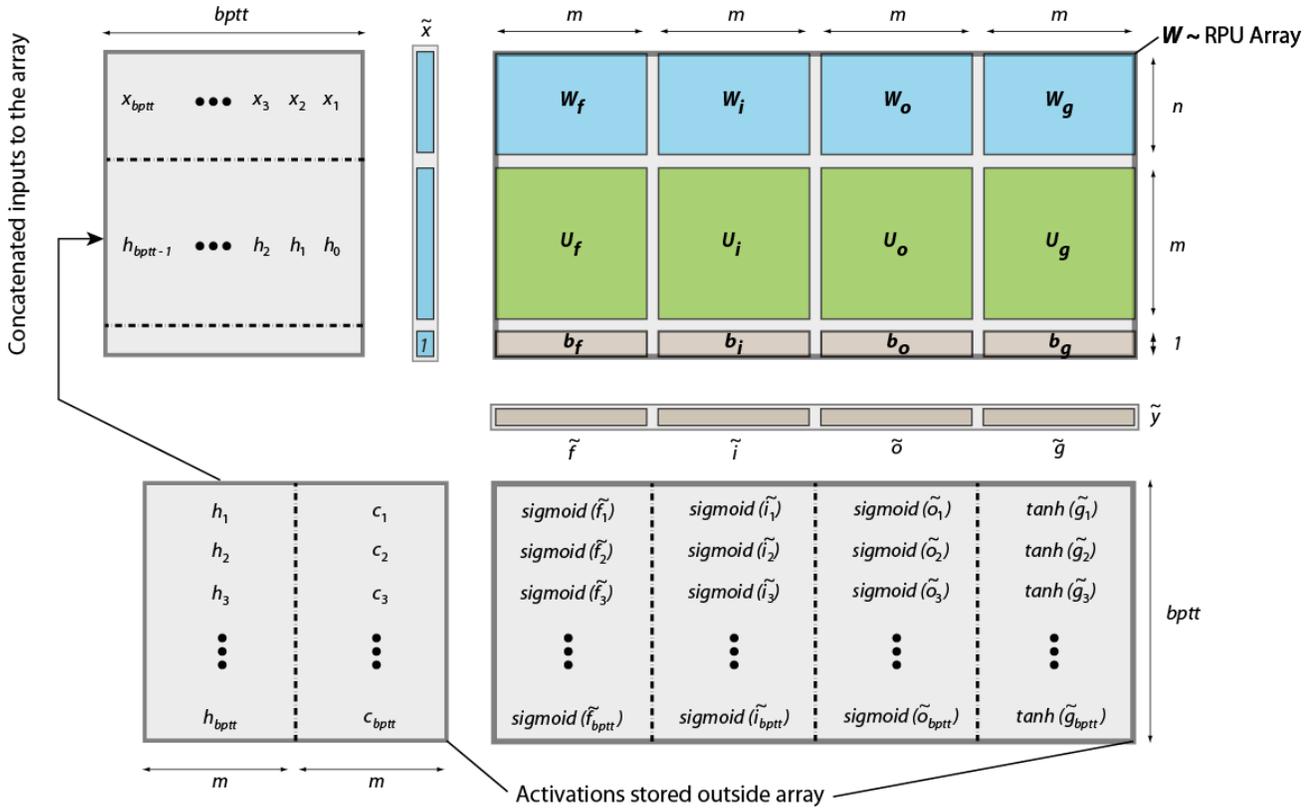

**Figure 2.** Schematics of an LSTM block mapped to an RPU array. The input vectors to the RPU array and the output vectors from the array are shown for the forward pass only. All activations are calculated and stored outside the array by digital circuits.

**RESULTS**

In order to test the validity our approach to map LSTMs to RPUs, we train LSTM networks similar to those described in [13], composed of 1 or 2 stacked LSTM blocks, with different number of hidden vector sizes of 64, 128, 256 or 512 on two datasets: Leo Tolstoy's War and Peace (WP) novel and Linux Kernel (LK) consisting of 3,258,246 and 6,448,461 characters, respectively. We split the data into training and test sets as 2,933,246 and 325,000 characters for WP and 6,111,421 and 337,040 characters for LK where each dataset, respectively, have a total vocabulary of 87 and 101 characters. Throughout the paper we use the following naming convention consisting of the network block, stacking, hidden vector length and the dataset. For instance LSTM2-512-WP is a 2 stacked LSTM network with a hidden vector size of 512 trained on the WP dataset. Following the mapping described above for LSTM2-512-WP we use 3 different arrays with sizes $2048 \times (512 + 87 + 1)$ and $2048 \times (512 + 512 + 1)$ for the two LSTM blocks and a third array of size $87 \times (512 + 1)$ for the last fully connect layer before $softmax$ activation. We note that the total number of trainable parameters for the largest networks trained here are about 3.4M and the total number of multiplication and summation operations that needs to be performed during a single training epoch is about $10^{14}$. These large number of operation makes these simulations about 1500x more challenging than training the MNIST dataset on a fully connected network [22].

**Optimization Approach**

It is critical to perform training simulations that can be supported by the real RPU array hardware. We note that the operations performed on the RPU array during the update cycle are all parallel but only supports a form

$$\boldsymbol{W} \leftarrow \boldsymbol{W} + \eta(\tilde{\delta}\tilde{x}^T) \tag{7}$$

which is an outer product of two vectors and a weight update combined into a single operation. This form is consistent with the simple SGD rank-1 update but any variant of a SGD such as RMSProp, Adagard, momentum, etc., all require the calculation of the gradient values first and then updating the weight value using some history dependent parameter per weight that is a function of previous weight values and/or gradients. In its most general form these operations can be written as a two-step process

$$\Delta \boldsymbol{W} = \tilde{\delta}\tilde{x}^T \tag{8}$$
$$\boldsymbol{W} \leftarrow \boldsymbol{W} + \eta(\boldsymbol{W}_{pre}, \Delta \boldsymbol{W}_{pre})\, \Delta \boldsymbol{W} \tag{9}$$

where the first operation is the gradient calculation and the second is the weight update. On a digital hardware those calculations may be insignificant and can easily be implemented by storing and updating one additional parameter per weight and do not increase the computational complexity. However, for RPU arrays such an extra operation will break array parallelism as the update cannot be performed at constant time. The calculations of the gradients given by Eq. (8) can still be performed on a separate array with $O(1)$ time complexity, however, Eq. (9) can only be implemented column-wise serially with $O(m)$ time complexity and therefore violates the array parallelism.

In order not to violate parallel array operations, in our simulations training is performed using only simple SGD. Additionally, mini-batch size of unity, fixed learning rate and time unrolling steps $bptt$ of 100 is used. Since these settings are slightly different from what is used in [13] (such as RMSProp with mini-batch size of 100), we first validated our training by performing simulations using high precision floating point (FP) numbers/operations and we tried various learning rates, $\eta$, with different amount of dropout rates, $p$, for each model individually. We note that the dropout term is only introduced for non-recurrent connections following the guidelines from [33] and is consistent with [13]. Figure 3 shows the best baseline-FP results for various LSTM2-WP models with different hidden vector sizes at the corresponding learning rate and dropout rates. For each model the test cross-entropy loss is on par or slightly better than the value reported by [13] and therefore validates our simple SGD training approach.

**RPU Baseline Model**

The various RPU device imperfections and their effect on the training accuracy is tested for a fully connected [22] and a convolutional neural network [29] on the MNIST dataset. Although the same device specifications were sufficient to train both networks successfully, input/output signal normalization/renormalizations were needed for successful training of CNNs. Here in our simulations we start with a baseline model that has identical device parameters and signal normalization techniques described for CNNs [29].

The RPU-baseline model uses the stochastic update scheme [22], where the length of the stochastic stream is $BL = 10$. The gain factors $C_x$ and $C_\delta$ used for determining the pulse probability during the stochastic translation for the columns and the rows are scaled properly $\left(C_x = C_\delta = \sqrt{\eta/(BL\,\Delta w_{min})}\right)$ to give the desired learning rate, $\eta$, used for training the model; and $\Delta w_{min}$ is the average incremental change in the weight value due to a single coincidence event. Although the average value for $\Delta w_{min}$ is set to 0.001, in order to capture device imperfections, $\Delta w_{min}$ is assumed to have cycle-to-cycle and device-to-device variations of 30%. Possible asymmetry in the weight updates are taken into account by using separate $\Delta w_{min}^+$ for the positive updates and $\Delta w_{min}^-$ for the negative updates for each RPU device. The average value of the ratio $\Delta w_{min}^+/\Delta w_{min}^-$ among all devices is assumed to be unity but with a 2% device-to-device variation. The bounds on the weights values, $|w_{ij}|$, is set to be 0.6 on average with a 30% device-to-device variation. For any real hardware implementations of RPU arrays results of the vector matrix multiplications will be noisy and this noise is considered by introducing an additional Gaussian noise, with zero mean and standard deviation of $\sigma = 0.06$. Furthermore, the results of the vector-matrix multiplications are bounded to a value of $|\alpha| = 12$ to account for signal saturation. The input signals are assumed to be between [-1, 1] with a 5-bit input resolution, whereas the outputs are quantized considering a 9-bit ADC. Although the inputs signals going into the array and the outputs signals coming from the arrays are bounded, we use the noise management and bound management techniques described in [29]. In particular, the inputs/outputs are normalized/renormalized using to the absolute maximum value of the elements of vectors $\tilde{x}$ or $\tilde{\delta}$ during the forward and backward pass, respectively. These normalizations are crucial not only because of small backpropagating error signals as discussed in [29] but also during forward propagation, because values in $\tilde{x}$ can go beyond unity due to the dropout term used: Note that during training time when dropping a random fraction $p$ of activations, the remaining are scaled with $1/(1-p)$ and therefore the input signals go beyond unity.

The test error of this RPU-baseline simulations on various LSTM2-WP models with different hidden vector sizes are shown in Figure 3 as black curves. Each model uses the same learning and dropout rates as for the corresponding FP-baseline model. In contrast to the behavior observed for the FP-baseline models, test errors of the RPU-baseline models increase and become noisier when the size of the network is enlarged. This is a very disappointing result and if not addressed, may limit the application space of analog device arrays to only a very small network sizes.

In order to identify the main cause of this problem, we performed training at various training conditions. For the models that are trained with a larger input signal resolution of 7-bits (but otherwise identical device and system properties), as shown by red curves in Figure 3, the test error follows a trend much more similar to the FP-baseline model. Although, there remains some offset between the FP-baselines and the RPU-models trained with 7-bit input resolution, offsets tend to get smaller and RPU-models improve in performance as the number hidden vector size (or parameters) increases. These results show that the undesired behavior observed for the black curves (RPU baseline) are solely due to the limited input signal resolution of 5-bits.

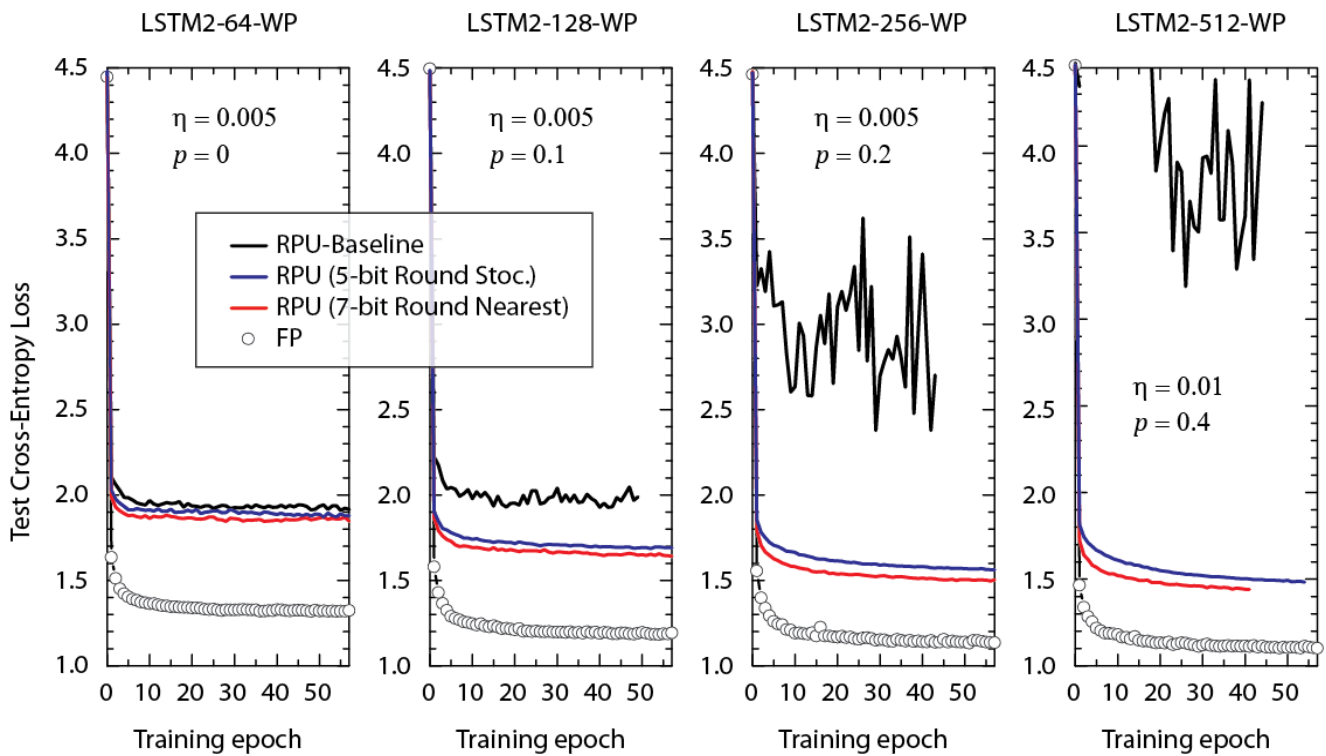

**Figure 3.** Test cross-entropy loss of two stacked LSTM networks (at different hidden vector sizes) trained on the WP dataset. Open white circles correspond to the model where the training is performed using floating point (FP) numbers. Lines with different colors correspond to RPU-baseline models using different input signal resolutions and rounding schemes as given by the legend. Same dropout probability, $p$, and the learning rate, $\eta$, are used for the FP and RPU models for each network size. For the sake of comparison we did not optimize these parameters with respect to the RPU model.

**Stochastic Rounding for Input Signals**

It is clear that the limited input signal resolution needs to be addressed for successful application of the RPU approach on large networks, however, increasing the input signal resolution comes with a cost of increased peripheral circuit complexity or computation time. For instance, for time encoded signals, increasing the input resolution from 5-bits to 7-bits increases signal duration by a factor 4 for the largest input, and therefore increases the computation (integration) time during forward and backward passes. Alternatively, for a fixed integration time, 7-bit inputs require 4 times faster clock rates during signal generation and therefore it may not be possible given the limitations due to signal filtering and clock rates. Using voltage height controlled inputs also comes with a cost, as more voltage levels need to be generated by the peripheral circuits which again increases the circuit complexity.

Here, we propose to use a stochastic rounding scheme (instead of rounding to nearest neighbor) as a cost effective solution for the input signals while still keeping signal resolution at 5-bits. It is already shown that stochastic rounding helps during DNN training when used at different stages of approximate computing with reduced precision in the digital space [34]. However, to prove the effectiveness of stochastic rounding for training RPU arrays, we performed simulations using the same RPU-baseline model with 5-bit input resolution but instead with the stochastic rounding scheme. As shown by the blue curves in Figure 3, stochastic rounding at 5-bit input resolution give results almost identical to round-to-nearest-neighbor scheme at 7-bits and therefore it can be a viable approach for real hardware implementations. The overhead of using stochastic rounding instead of rounding to nearest neighbor is very small [34] and it can be realized by specifically designing additional hardware residing in the digital blocks that moves data between RPU arrays and NLF units. Although our simulations do not guarantee that the 5-bit input resolution can be universally applicable for even larger networks, it is clear that using stochastic rounding saves a couple of bits during input signaling and hence improves the overall performance of the RPU arrays.

**Effect of Dropout**

It is known that successful applications of large neural networks require good regularization and dropout [35] is one of the most powerful regularization methods. Indeed, we also use dropout in our training and larger dropout rates are needed for the best performance as the network size gets larger as shown by the FP-models in Figure 3. To highlight the importance of dropout, we show the test results of LSTM2-512-WP, the largest network of interest, trained at different dropout probabilities in Figure 4(a). It is clear that small dropout rates ($p < 0.4$) cause the networks to overfit as the test errors start to increase after a certain amount training. Only the cases with a 40% dropout rate or higher show a consistent down trend in the test error and hence eliminates the overfitting problem. However, increasing the dropout rate arbitrarily beyond 40% is not beneficial either (data not shown) and the best generalization results are observed at about 40% dropout rates for LSTM2-512-WP when trained with floating point numbers.

In order to test the effect of dropout for a realistic hardware implementations of RPU arrays, we performed training using the RPU-baseline model at 7-bit input resolution and varied the dropout rates, as shown by Figure 4(b). In contrast to the results obtained by the FP-models, even when the dropout is completely eliminated we did not observe overfitting to be a problem. In addition, the best performance is obtained for dropout rates at around 10-20%, which is smaller than the optimum value used for FP-models at 40%. These results suggest that for a realistic implementation of RPU arrays the training may not require such a strong regularization term or the same amount may be non-optimal as there exists many sources of noise and stochasticity coming from the hardware. However, it is also important to realize that the effect of the dropout is much smaller for all RPU models; and even with the optimum dropout rates we consistently observe an offset between the RPU-models and FP-models for all LSTM sizes (data for smaller networks are not shown).

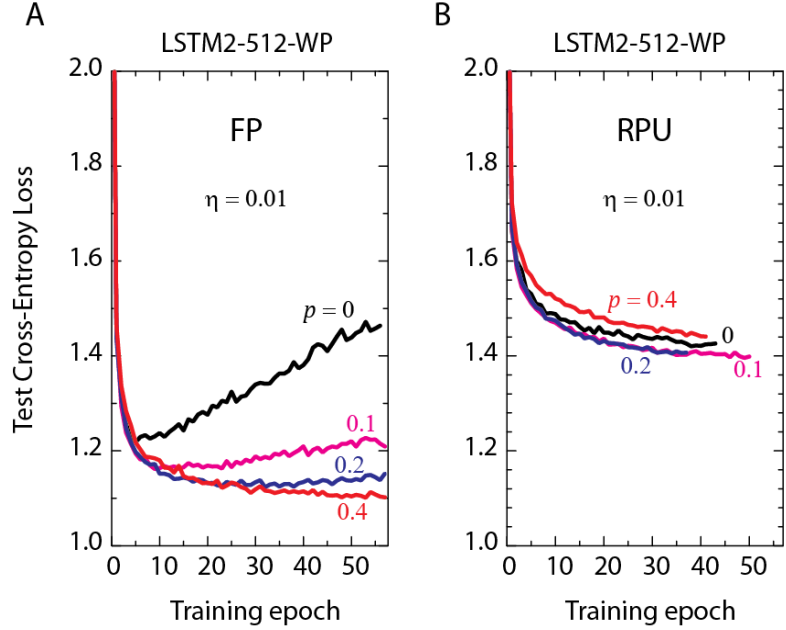

**Figure 4.** Test cross-entropy loss of two stacked LSTM networks with a hidden vector size of 512 trained on WP dataset. Lines with different colors correspond to **(A)** the model trained using floating point (FP) numbers **(B)** the RPU-baseline models using 7-bit input signal resolution, at different dropout probabilities, $p$.

**Effect of Device Variations, Asymmetry and Number of States**

To understand the main cause of the offset observed between FP-baseline and RPU models we performed training using a range of RPU models. In each we selectively eliminate device imperfections to evaluate their influence on training performance. The summary of these training results on LSTM2-WP with different hidden vectors sizes are shown in Figure 5. The green curves in Figure 5 correspond to the models where device-to-device and cycle-to-cycle variations in the parameters $\Delta w_{min}$ and $|w_{ij}|$ are completely eliminated from their original values at 30%. Interestingly, eliminating of all this variability from the model does not improve the network performance compared to the RPU-baseline models as shown by red curves. Similarly, the cyan curves corresponding to RPU models with a larger number of states (4x more compared to baseline) also show test errors that are almost indistinguishable from the RPU-baseline model. Only the blue curves corresponding to RPU models without any device-to-device variation in the asymmetry parameter ($\Delta w_{min}^+/\Delta w_{min}^-$) show an improvement and test errors get closer to the value achieved by the FP-baselines (shown by open circles). These results suggest that the number one factor that is limiting the performance of these networks is the asymmetry parameter and even a slight asymmetry term with only a 2% percent device-to-device variation is sufficient to be harmful. Only after the elimination of the asymmetry term, increasing the number of states further enhances the network performance as shown by magenta curves. We note that all RPU models shown in Figure 5 are simulated using a 7-bit input resolution in order not to introduce additional artifacts due to limited input resolution; and each model used the same dropout and learning rate that the FP-baseline is trained with.

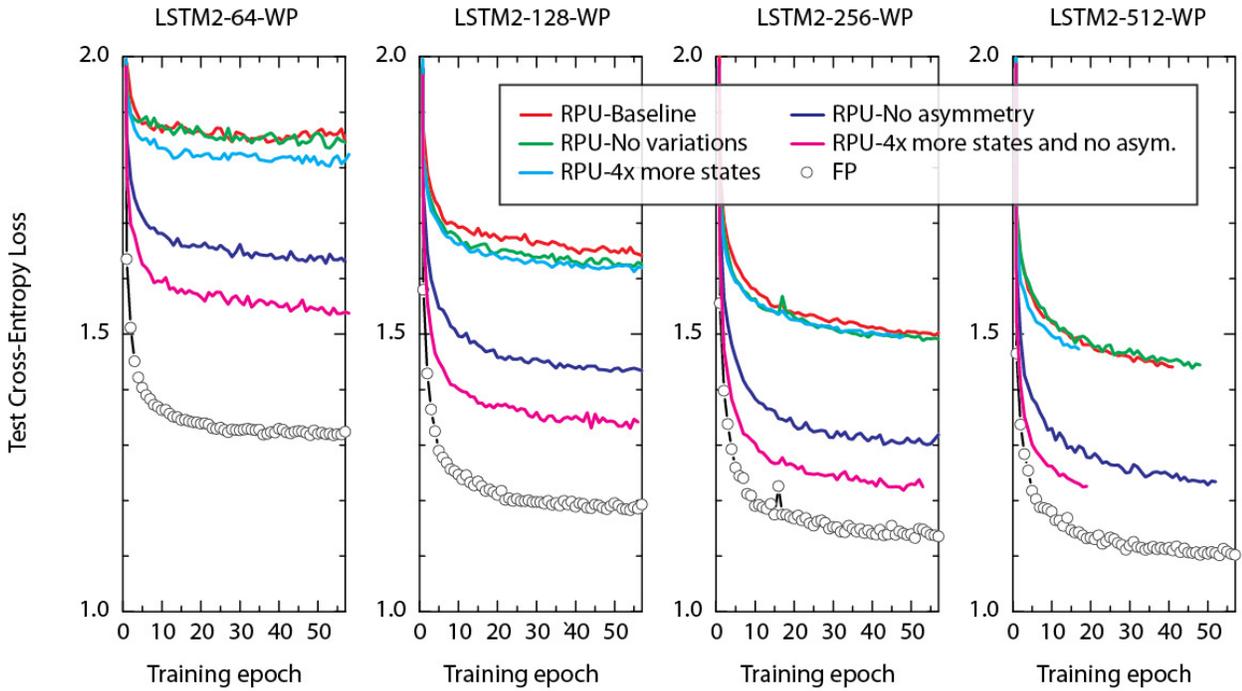

**Figure 5.** Test cross-entropy loss of two stacked LSTM networks (at different hidden vector sizes) trained on WP dataset. Open white circles correspond to the model where training is performed using floating point (FP) numbers. Simulation of RPU-baseline models are shown by red curves. Lines with different colors correspond to RPU models but for each model a set of device imperfections are selectively eliminated compared to the RPU-baseline model. Green curves: device-to-device and cycle-to-cycle variations in the parameters $\Delta w_{min}$ and $|w_{ij}|$ are completely eliminated. Cyan curves: the total number of states is increased by 4x. Blue curves: device-to-device variation in the asymmetry parameter is eliminated. Magenta curves: as for blue curves, but additionally the number of states are increased by 4x. All RPU models are trained with 7-bit input signal resolution with round-to-nearest-neighbor scheme. Since the simulation run times were limited to 7 days, some curves stop early before reaching 50+ epochs.

## DISCUSSION AND CONCLUSIONS

The application of the RPU device concept for training LSTM blocks is shown to be very similar to the training of fully connected layers. A single vector operation performed on the RPU array computes all of the linear transformations needed for a single time step in parallel regardless of the LSTM block size. We assume that all other non-linear operations are performed outside the array by using programmable digital circuits and that the results are sent back to the same RPU array for the next step of the calculations. This realizes the weight sharing that happens between different time steps. We note that these programmable digital blocks control the signal flow and also perform different types of computations; hence it becomes very easy to implement other kinds of recurrent networks on the same hardware. For instance gated recurrent units (GRU) [36], dilated RNNs [37] or other more complex RNN architectures [38] can be mapped to RPU arrays in a similar fashion by simply changing the computations performed on the digital circuits. We emphasize that the concepts described here are more general and can be applied to map other more complex neural networks including a mixture of recurrent, convolutional and fully connected layers by simply reprogramming digital blocks that compute the non-linear functions and control the signal flow.

The operations performed on the arrays are identical for different network architectures including fully connected, convolutional or recurrent networks. However, it is not obvious that the same device constraints derived from a small fully connected network can be generalized to give competitive training accuracies for larger and complex networks on larger datasets. The LSTM networks studied here is an attempt to test the effect of different hardware noise and device variations on the training performance in a much more challenging task. We note that the array sizes used for the LSTM2-512-WP model are $(2048 \times 600)$, $(2048 \times 1029)$ and $(87 \times 513)$ and these arrays are much larger than the ones used for the fully connected network studied in [22] with sizes of $(256 \times 785)$, $(128 \times 257)$ and $(10 \times 129)$. In addition, the training sequence is consisting of about 3M characters for the WP dataset compared to 60K training images in the case of MNIST dataset. The combination of larger array sizes and more training examples makes the training of these LSTM networks about 1000x (1500x for LK dataset) more challenging than training the MNIST dataset on the aforementioned fully connected network. Interestingly, the 2% variation in the asymmetry term that was sufficient to train the fully connected network at the level of floating point model accuracy is shown to be not sufficient for these LSTM networks. This result suggests that the asymmetry parameter becomes increasingly more critical for larger scale networks and it may require special attention during hardware design and development.

The performance benefits of the RPU approach for LSTM networks can be calculated using the design considerations described in [22]. For an LSTM block, computation steps dependent on the computations in previous time steps and additionally on computations in the previous LSTM blocks if stacked LSTM networks are used. Therefore, a pipelined microarchitecture design is required to utilize multiple RPU arrays that can perform concurrent computations corresponding to the different computation steps of the data. Assuming a fully pipelined architecture and the LSTM2-512-WP model, there would be a total of 3.4M RPU devices active at any given time residing on 3 different arrays. Using a measurement (cycle) time of roughly $t_{meas} = 80\ ns$ [22] for each forward, backward and update cycles, we can estimate the total RPU accelerator chip performance using the below simple formula

$$Throughput = \frac{2 \times Total\_RPU\_Count}{t_{meas}}\ Ops/s \qquad (10)$$

where the factor of 2 comes from the multiplication and the summation operations performed on each RPU device. This yields a throughput of $85\ TeraOps/s$ for the LSTM2-512-WP model. This is already significantly higher than the peak single precision throughput of an NVIDIA Tesla P100 at about $10\ TeraOps/s$. Thus the performance benefits of the RPU approach becomes already apparent for the sizes of the LSTMs investigated here, and once it is applied to much larger problems with a total number of RPU devices reaching billions, throughput of an RPU accelerator chip exceeds the throughput of todays' advanced GPUs and accelerators by more than 1000x.

In order for RPUs to be a competitive technology, however, the symmetry requirement of the weight update needs to be addressed. Accomplishing such symmetrically switching analog devices as needed is a difficult task. Besides material engineering, circuit assisted solutions combined with algorithmic modifications might, conceivably, relax the material requirements. One example of an almost perfectly symmetric RPU is demonstrated by designing analog CMOS [39] [40] (so called CMOS-RPU) that performs the updates using a current source and sink circuitry and stores the weight as charge on a capacitor. In this design, it is shown that symmetry is achieved by properly balancing the current source and sink that incrementally change the stored charge on the capacitor. Device leakage, device mismatch and charge retention on the capacitor are critical components for the scalability to larger networks. Functionality of this RPU concept comes at the cost of significant circuit overhead. In contrast to the CMOS-RPU approach, there are device options available that may be used to realize the RPU concept. One noteworthy device concept is the so called LISTA device [25] that shows significantly more symmetric behavior if a current pulsing scheme is used. However, this current pulsing scheme would also require a current source and sink circuitry similar to the ones used in the CMOS-RPU design. A simple constant voltage pulsing scheme is difficult to realize for the demonstrated LISTA devices in an array configuration due to the built-in voltage which depends on the individual weight state of each node. By properly selecting the materials used in the device stack this built-in voltage problem can be mitigated and it is an interesting research direction for realizing a symmetric RPU concept. Finally, we note that PCM devices [27] [23] are promising candidates to realize the RPU concept. PCM elements change their conductance gradually at one polarity (SET) and very abruptly at the opposite polarity (RESET). Therefore, the weight is encoded in a pair of PCM elements that operate in SET mode in a differential configuration. Non-linearities and conductance saturation are detractors for optimal performance. However, using appropriate CMOS circuit elements these detractors can be overcome and provide a possible solution for deep learning [31].

It is clear that a global asymmetry term, uniform among all devices, can be fixed easily by the supporting peripheral circuits using different voltage pulses for up and down changes for the whole array without requiring a serial access to each device. However, if there is a slight device-to-device variation that causes a local asymmetry term such a compensation is not possible without leaving the parallel operation of the array. Given that these arrays would be fully utilized and always busy in a pipelined design to get the most performance benefits, any kind of interruption to the parallel operations may become too costly no matter how infrequent the interruption is. Therefore, the area, power and especially the time cost of these engineering solutions need to be sized properly as it may significantly reduce the benefits of using analog arrays for DNN training.

In summary, we believe that the RPU concept is a very promising candidate to accelerate the training of a range of complex deep neural networks, however, its success strongly depends on realizing a cross-point that can change its state in a symmetrical fashion. Once the symmetry problem is overcome, the RPU concept can provide unprecedented acceleration factors reaching 10,000x compared to the digital counterparts [22]. For a highly optimized digital hardware one can think of fitting tens of thousands of multiplication and summation units on a single chip. However, even these numbers look miniscule when compared to an RPU approach, as a single RPU array consisting of 4096x4096 cross-points can perform 16 million multiplication and summation operations all in parallel in the analog domain by using only a fraction of the chip area. Using multiple arrays simultaneously would make the throughput of analog accelerator chip even more impressive reaching 3-4 orders of magnitude larger than the digital only solutions. Therefore, large problems of interest for business applications that currently require days of training on multiple digital hardware can take only minutes using a single RPU based analog accelerators.